\documentclass[sigconf, nonacm, authorversion]{aamas}

\usepackage{balance} 
\usepackage{amsmath,amsthm}
\usepackage{algorithm}
\usepackage{algpseudocode}
\usepackage{graphicx}
\usepackage{xcolor}
\usepackage{multirow}
\usepackage{float}
\usepackage{tikz}
\usepackage{framed}
\usepackage{hyperref}

\usepackage{colortbl}
\definecolor{lightblue}{RGB}{173,216,230}

\usepackage{soul}
\sethlcolor{blue!15}

\usepackage{dsfont}

\usepackage{pgf}

\everymath=\expandafter{\the\everymath\displaystyle}

\newcommand{\callsign}{\textsc{Streetwise}}

\newcommand{\exop}{\mathcal{E}}
\newcommand*\circledfill[1]{\tikz[baseline=(char.base)]{
            \node[shape=circle,fill=gray,text=white,line width=0pt,draw,inner sep=2pt] (char) {#1};}}

\acmSubmissionID{807}

\title{\callsign~ Agents: Empowering Offline RL Policies to Outsmart Exogenous Stochastic Disturbances in RTC}

\author{Aditya Soni, Mayukh Das, Anjaly Parayil, Supriyo Ghosh, Shivam Shandilya, Ching-An Cheng, Vishak Gopal, Sami Khairy, Gabriel Mittag, Yasaman Hosseinkashi, Chetan Bansal}
\affiliation{
  \institution{Microsoft}
  \city{}
  \country{}}
\email{}

\begin{document}
\pagestyle{plain}
\begin{abstract}
The difficulty of exploring and training online on real production systems limits the scope of real-time online data/feedback-driven decision making. The most feasible approach is to adopt offline reinforcement learning from limited trajectory samples.
However, after deployment, such policies fail due to exogenous factors that temporarily or permanently disturb/alter the transition distribution of the assumed decision process structure induced by offline samples. This results in critical policy failures and generalization errors in sensitive domains like Real-Time Communication (RTC). We solve this crucial problem of identifying robust actions in presence of domain shifts due to unseen exogenous stochastic factors in the wild.
As it is impossible to learn generalized offline policies within the support of offline data that are robust to these unseen exogenous disturbances, we propose a novel post-deployment shaping of policies (\callsign), conditioned on real-time characterization of out-of-distribution sub-spaces. This leads to robust actions in bandwidth estimation (BWE) of network bottlenecks in RTC and in standard benchmarks. Our extensive experimental results on BWE and other standard offline RL benchmark environments demonstrate a significant improvement ($\approx$ 18\% on some scenarios) in final returns wrt. end-user metrics over state-of-the-art baselines. 
 
\end{abstract}

\maketitle

\section{Introduction}



In most decision-making problems in real safety-critical and sensitive domains such as Real Time Communication (RTC), Robot Navigation, AI-assisted medical procedures and so on, online data collection and learning can be impractical, or even impossible, due to the associated costs or safety concerns \cite{dulac2021challenges}. In such domains exploration and getting feedback on production environments is not feasible. A more practical approach is to employ scalable, data-driven methods that can learn policies from existing data. Offline reinforcement learning (Offline RL) allows for controlled and deliberate learning from previously collected data samples \cite{levine2020offlinereinforcementlearningtutorial}. 

\textit{Promises and Limitations of Offline RL:} The generalizability and performance of offline policies heavily depend on the support and coverage of the offline dataset. There is active research on improving the generalization offline RL policies via regularization \cite{kumar2019stabilizing, siegel2020keep,brandfonbrener2021offline} and on additional data augmentation to improve the coverage or on ensuring survival instinct \cite{li2023survival}. However, many of these approaches fail to account for domain shift after deployment of learned policies.
Domain shift can be characterized in many ways but, perhaps, one of the most critical ones occur when exogenous stochastic signals that were not present during training are encountered in real deployed environments, leading to unexpected and potentially undesirable outcomes. This is particularly common in domains such as real-time communication (RTC), where network dynamics can change temporarily or permanently. 


\begin{figure}[t]
    \scalebox{0.3}{\input{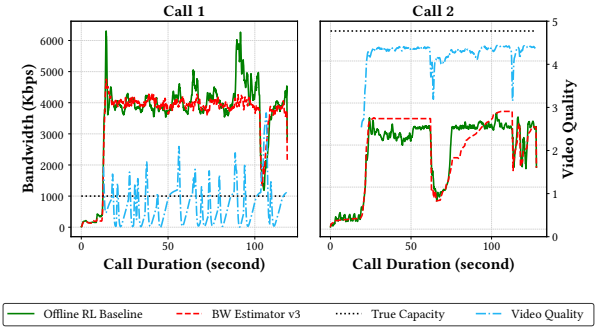}}
    \caption{Example calls with exogenous disturbances leading to sub-optimal predictions by deployed policies in Bandwidth Estimation domain}
    \vspace{-1.5em}
  \label{fig:bwertc}
\end{figure}

\textit{Domain and Nuances:} Real-Time Communication and Control problems (our domain of interest in the context of this work) involve optimizing audio/video calls and media streaming applications over networks. One of the most crucial problems in RTC is Bandwidth Estimation (BWE) of the bottleneck connections in a network \cite{fang2019reinforcement,bentaleb2022bob}.  The fidelity of this estimation controls is critical since it affects the target bit-rate and, hence, is key to effectiveness of other call/streaming quality optimization models and frameworks. Traditionally, statistical non-linear sequential regressors such as Unscented Kalman Filters (UKF) \cite{wan2000unscented, fang2019reinforcementlearningbandwidthestimation} have been used for this task. While such frameworks are effective for certain types of network dynamics, they are brittle and do not adapt well to the dynamic shifts in the network bottlenecks. Reinforcement Learning (RL)-driven solutions for BWE has become increasingly popular  \cite{fang2019reinforcementlearningbandwidthestimation,SeCBAD} and have shown drastic improvements over statistical models and sequential regressors. However, despite the promise, online RL approaches are infeasible in real scenarios as it is not viable to train on production systems. Hence offline RL is one of the most appropriate choices for designing bandwidth predictors. 

Learning robust offline RL policies for BWE in RTC is one of the most tricky continuous control problems. In this domain, we can easily observe presence of unseen exogenous stochastic signals that disturb the network dynamics during real audio/video calls. Figure~\ref{fig:bwertc} shows 2 such example calls where policies are over-predicting bandwidth (left) at runtime, diminishing the returns (Mean Opinion Score - MOS), possibly due to some unknown exogenous signal that affected the dynamics in some of the dimensions, such as "packet receive rate", that spuriously made the policy perceive underutilized bandwidth. The other example (right) shows sparse intermittent disturbances in which the statistical models (BW Estimator V3 - such as UKF or WebRTC based frameworks \cite{fang2019reinforcement, webrtc}) as well as the naive offline policy takes time to recover. While bandwidth estimation is one of the most notorious example domains where post-deployment domain shift from exogenous unknown signals is common, this phenomenon is not uncommon in other sequential decision-making problems, ranging from autonomous navigation and control to systems optimization 
\cite{wang2024research,wu2023human}.

\textit{The problem:} While the effect of exogenous signals on the decision process can be arbitrary, in the context of this work, we focus on the most dominant and complex scenario, where it intermittently or permanently changes the transition dynamics and sometimes even the state space from what is observed in offline  samples. This creates Out-Of-Distribution (OOD) regions or subspaces. It is theoretically and practically infeasible to train generalized policies, within the support of the offline data, that are robust to such OOD regions in production. Several existing works do address the slightly different distributional shifts between the dataset and the learnt policy, and consequent over-estimation of the value of unseen state-actions by improving the estimation of the value function \cite{kostrikov2022offline, cql2020aviral, garg2023extreme, pmlr-v162-cheng22b} and behaviour-regularizing the policy gradient (e.g., TD3+BC) \cite{fujimoto2021minimalist}. Some also employ entropy based uncertainty estimation to quantify distributional noise and then learn value/policy ensembles to address that \cite{an2021uncertainty}. However, such arguably robust policies are not equipped to handle unseen exogenous stochastic disturbances to the transition dynamics in production post deployment, especially in sensitive and critical problems like BWE in RTC.

\textit{Our Contribution}: Can agents be smart enough to handle any unseen uncharacterized exogenous disturbances? Our proposed novel \callsign~agent is a significant step in that direction. Instead of trying to learn a generalized robust policy applicable to any nature of domain shift/OOD, which is impossible within the support of the limited offline samples, it shapes/perturbs the deployed offline policy at real-time post deployment. The shaping is conditioned on real-time characterization and quantification of OOD spaces from live observations in production environments. The nomenclature \callsign\footnote{Code at \href{https://anonymous.4open.science/r/PDS-C53E/}{https://anonymous.4open.science/r/PDS-C53E/}} draws on the analogy of people who are streetwise, i.e. they can handle any unknown possibly tricky situation out in the world. We show significant average gains ($\sim 15\%$ compared to backbone offline policy and $\sim 6\%$ compared to SOTA perturbation methods on lossy network profiles having arbitrary stochastic disturbances) in final returns in context of the BWE for near-real emulated calls. We also highlight that on stable networks where the backbone offline policy is already optimal, \callsign's performance is never worse than the backbone policy's performance ensuring safety. 
We make the following contributions: \begin{enumerate}
    \item We propose a novel \callsign~agent framework that can help deployed offline policies outsmart/handle OOD scenarios triggered by exogenous process. 
    \item We adapt this successfully to the BWE problem in RTC, showing significant gains in call/streaming quality, a significant step towards high fidelity bandwidth estimation and control.
    \item We also show empirically that our proposed post-deployment shaping can be successfully adapted to any continuous control domain beyond our target application context of RTC. We verify this with several benchmark D4RL/Mujoco tasks. 
\end{enumerate}
The rest of the paper is structured as follows -- We first discuss related literature in context of data-driven BWE, Offline Reinforcement as well as OOD mitigation in traditional and sequential prediction problems  (Section~\ref{sec:related}). We then define the problem setting and formalism and the notations (Section~\ref{sec:setting}). In Section~\ref{sec:main} we describe our \callsign~agent framework, its architecture and  algorithmic details. Finally, in Section~\ref{sec:eval} we present and discuss the performance of our approach on BWE task on near-real emulated calls across varied types of network profiles. We also highlight how \callsign~ generalizes to benchmark tasks in noise induced MuJoCo environments.



\section{Background and Related Work}
\label{sec:related}
\subsection{BWE in RTC}
Existing works on bandwidth estimation in RTC employ rule-based algorithms \cite{45646, gaetano, scavenger, cellular, copa} and learning-based methods \cite{pragmatic, onRL, loki} for bitrate adaptation. Some works design hybrid methods \cite{Wang2021AHR} that combine heuristic based algorithms with an RL agent to tune the bandwidth estimate of the heuristic congestion control scheme. 


Early investigations on deep RL for BWE use Proximal Policy Optimization (PPO)\cite{schulman2017proximalpolicyoptimizationalgorithms} to train policies on a simulator, with the goal of maximizing user's quality of experience (QoE) \cite{fang2019reinforcementlearningbandwidthestimation}. But sim-to-real generalization issues were quite significant which motivated several later research directions such as imitation learning \cite{gottipati2024offlineonlinelearningpersonalized} and offline meta-learning \cite{gottipati2024balancinggeneralizationspecializationoffline} to learn more specialized policies.

We focus and highlight the offline RL setting, since we cannot train BWE policies via online interaction with the environment. We learn from a static dataset that consists of call logs collected from multiple behaviour policies that employ rule-based methods and consequently na\"ive offline RL policies cannot handle OOD transition dynamics problem induced by an exogenous process acting on the environment post-deployment.

\subsection{Offline RL: Promises and pitfalls}
In many real-world problems, it is infeasible to train policies online because of resource constraints. Offline reinforcement learning \cite{levine2020offlinereinforcementlearningtutorial} allows training policies from previously collected datasets without any online interactions. Offline RL, however, comes with the problem of distributional shifts between the dataset and the learnt policy, which can lead the agent to overestimate the value of unseen state-actions. Several works try and mitigate the problem by improving the estimation of the value function \cite{kostrikov2022offline, cql2020aviral, garg2023extreme, pmlr-v162-cheng22b} and behaviour-regularizing the policy gradient (e.g., TD3+BC) \cite{fujimoto2021minimalist}. 

We instead look at the problem of exogenous disturbances present in the environment post-deployment. Irrespective of the state coverage after training a policy offline, an exogenous process can disturb the transition dynamics of the environment, rendering the environment non-Markovian. Since this is an offline learning setup, the problem cannot be posed as a non-Markovian optimization task without the support of offline samples. We instead interpret this as a problem of encountering OOD state transitions.

\subsection{OOD Detection and Mitigation}



\textbf{OOD detection under poor offline data support}. A related setting to ours is building robust outlier detection methods in the absence of offline data support. Prior works that train predictive models augment OOD detectors by generating surrogate OOD samples, either by perturbing a learnt model \cite{wang2023outofdistribution}, or by using generative models \cite{du2023dream}. These methods, however, cannot be directly applied to the sequential decision making domain. In offline RL, we do not have access to the underlying transition dynamics $T= p(s'|s, a)$ and reward dynamics $r$ of the environment, and learning a model for the same is a non-trivial task.

\noindent\textbf{OOD in Reinforcement Learning}. Domain shifts in Markov Decision Processes (MDPs) are defined in various ways across literature. Several works define OOD as a change in the state-action distribution, and propose algorithms that can help the policy generalize to unseen state-action pairs. Some works formalize disturbances in MDPs and discuss how OOD can occur not just in the state-action distribution, but in the environment's transition dynamics as well \cite{Haider2021DomainSI, oodThesis}. UniMASK and MaskDP \cite{unimask, liu2023masked} use transformer architectures to model trajectory snippets and use a single model with different masking schemes to adapt to  multiple downstream tasks, such as detecting OOD by setting a threshold for the sequence reconstruction error, and generating actions. 
Another trajectory of research explore ensembles of learnt dynamics models \cite{oodProbDynModel} or value functions \cite{an2021uncertainty} to estimate epistemic uncertainty. These methods, however, are ineffective if the ensembles are not exposed different categories of stochastic noise and are also compute intensive and hence difficult to deploy in the RTC domain. 


\noindent\textbf{Non-stationarity  and Bayes-Adaptive RL}. Another way to address our problem setting is to use Bayes-optimal policies to deal with uncertainty. Ghosh et al. \cite{ghosh2022offline} show that acting Bayes-optimally under uncertainty involves solving an implicit Partially Observable Markov Decision Process (POMDP). However, solving a POMDP requires inferring a belief state from the environment and maintaining it, which is intractable in real-world problems. To design more suitable methods, recent works constrain the definition of these environments to that of Hidden-Parameter MDPs (HiP-MDP), where the true state is defined as the combination of the observation $s$ and hidden parameter $z$, i.e. $\hat{s} = (s, z)$, or Dynamic-Parameter MDPs (DP-MDP), which assume that the hidden parameter $z$ is fixed during an episode but evolves across episodes \cite{Ackermann2024OfflineRL, pmlr-v139-xie21c}.

Several works propose Bayes-Adaptive RL methods to solve HiP-MDPs, such as VariBAD (Zintgraf et al., 2020) \cite{varibad}, BOReL (Dorfman et al., 2021) \cite{dorfman2021borel} and ContraBAR (Choshen and Tamar, 2023)\cite{contrabar}. VariBAD is an online learning algorithm that conditions the policy on a belief over the current HiP, and infers the belief using a Variational Autoencoder (VAE). BOReL introduces two data collection and modification techniques to apply VariBAD's method to offline datasets: \textit{Reward Replaying}, which augments each transition $ (s, a, r, s^\prime)$ in the dataset with transitions $(s, a, R(s, a, z_i), s^\prime)$ for all HiPs $z_i \in Z$; \textit{Policy Replaying}, which generates trajectory on an HiP using behaviour policies conditioned on all other HiPs $z_i \neq z$. These techniques, however, are not applicable to our setting because we are learning from a dataset where the trajectories were collected using HiP-agnostic behaviour policies, and we do not have access to the reward function to augment the dataset. Further, the majority of the experiments by Dorfman et al. (2021) \cite{dorfman2021borel} and Choshen and Tamar (2023) \cite{contrabar} focus on settings with varying reward functions, while our setting focuses on settings with varying transition functions.

Perhaps the closest settings to ours are studied in Chen et al., \cite{SeCBAD} and Ackermann et al., \cite{Ackermann2024OfflineRL}. Chen et al., \cite{SeCBAD} explore making policies robust to changing contexts across call segments in RTC, but their work only focuses on the online RL setting. Ackermann et al., \cite{Ackermann2024OfflineRL} study non-stationarity under the constraints of the offline RL setting, but they make additional assumptions about the occurrence of non-stationarity in the dataset to interpret the setting as a DP-MDP, which does not hold for the RTC domain. 
Both of these works learn a latent variable or a context vector from the training data and condition the policy on it. Our problem is different from these because we operate in the context of training with limited offline data support with no representation of unknown exogenous disturbance, 
hampering the training of policies conditioned on a latent variable. Thus, we explore opportunities in post-deployment policy shaping to make the model adapt to exogenous disturbances.

\section{Problem Setting}
\label{sec:setting}

Given a Markov Decision Process, MDP, $\mathcal{M} = (\mathcal{S}, \mathcal{A}, R, T)$, with, continuous state and action spaces $\mathcal{S}$, and $\mathcal{A}$ resp., reward function $R = r(s, a)$ and transition dynamics $T= p(s^\prime|s, a)$ (assuming $\rho_0(s)$ be initial state distribution, and $\gamma$ the discount factor), Reinforcement Learning  aims to learn a policy $\pi(a|s)$ that maximizes the expected return $J(\pi) = \mathbb{E}_{\tau \sim p_M^\pi(\tau)} \left[ \sum_{t=0}^\infty \gamma^t r(s_t, a_t)\right]$, where $\tau = (s_0, a_0, s_1, a_1, ...)$ is a trajectory.  In offline RL, we learn a policy from a static dataset of transitions $\mathcal{D} = \{(s_i, a_i, r_i, s^\prime_i)\}_{i=1}^N$ collected by one or more behaviour policies $\pi_b(a|s)$.


In this work, we consider a setting in which exogenous stochastic process $\mathcal{E}$ can affect the real environment on which the agent is deployed after its policy has been learned from offline data. This forces the policy to infer on Out-Of-Distribution subspace. We formalize these OOD spaces as disturbances to the transition dynamics, and define the new dynamics $T_{\nu} = T_{(\mathcal{D})} + T_\exop$. So $T_{\exop}(s^\prime_\nu|s_\nu,a)\sim \exop$ where $s_\nu, s^\prime_\nu \in S_\nu = S\cup S_\exop$ and $S_\exop \sim \exop$. This effectively renders the post deployment environment non-markovian, but we cannot train a non-markovian policy within the support of $\mathcal{D}$

Note that it is theoretically and practically impossible to recover $\mathcal{E}$ during offline training, since $S_\exop,T_\exop \notin \mathcal{D}$. Now, let us assume there exists a globally optimal policy $\pi_\nu^* : \pi_\nu^* = \pi_{\mathcal{D}}^* \cup \pi^*_{\exop}$. Ideally our goal would be to learn $\pi^*_{\exop}(a|s\in S_\nu)$ but that is not learnable within the support of $\mathcal{D}$ since $(S_\exop \subset S_\nu)  \notin S$ and $T_\exop \notin D$

To address this challenge, our problem setting is the following,
\begin{leftbar}
\noindent\textit{\textbf{Given}: Exogenous stochastic disturbance $\exop$ in post deployment real environment and trained offline policy model $\pi^*_\mathcal{D}$\\
\textbf{Goal} is to design a shaping potential function $\phi(\cdot|\pi^*_\mathcal{D},\rho_\exop)$ to shape the action at run time, such that, 
\begin{equation}
    \label{eq:probsetting}
    a\sim \pi^*_\mathcal{D}(s\in S_\nu) \oplus k\sim \phi(\cdot|\pi^*_\mathcal{D},\rho_\exop) \approx a_\nu \sim \pi_\nu
\end{equation}
where $\rho_\exop$ is a measure in some metric space that quantifies the position, span and amplitude of unknown exogenous stochastic  and in particular and the distribution of this measure over the metric space maps to the structure of the unknown distribution in $\exop$:$\rho_\exop \mapsto \exop$. $k$ is just a value sampled from potential $\phi$.}
\end{leftbar}

While the state/action space designs may vary between are our target application domain BWE and others, the formalism of the exogenous disturbance and our problem setting is generic across any domain with similar problem space. 

\begin{figure*}[htbp]
    \centering
    \includegraphics[width=0.9\textwidth]{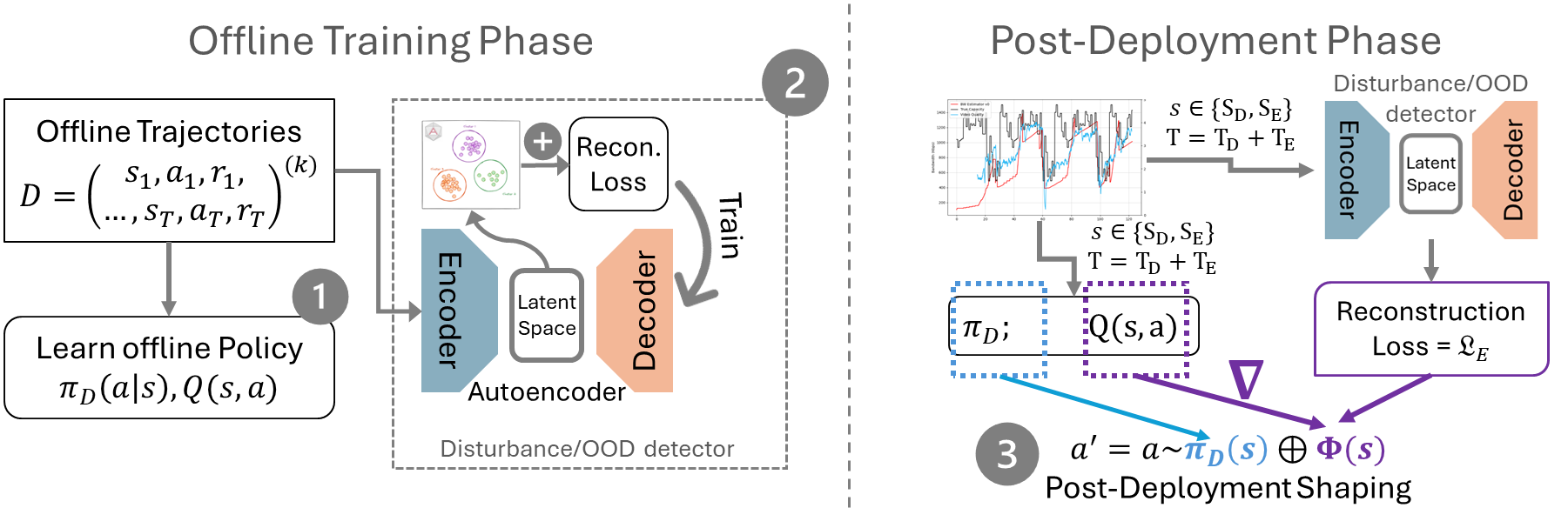}
    \caption{\callsign~agent architecture. \texttt{Left} shows the learning/training phase on limited offline samples, where Offline RL algorithm trains a policy $\pi_\mathcal{D}$ and also preserves the value function $Q_\mathcal{D}$. We also train an Autoencoder that will detect and quantify OOD after deployment. \texttt{Right} shows the post-deployment framework, where new given new observations from both offline and OOD transition distributions and state distributions employs the OOD/Disturbance detector to compute reconstruction loss as metric for the position, span and amplitude of OOD regions. It then shapes/perturbs the predicted action with a shaping potential $\phi$ based on the gradient of the Value function and the reconstruction loss}
    \label{fig:methodoverview}
\end{figure*}


\section{\callsign~Agents}
\label{sec:main}
We introduce our proposed \callsign~agent, 
which is a post-deployment policy shaping framework. While the term policy shaping usually applies to altering and refining policy with human-in-the-loop or knowledge guided signals during training \cite{griffith2013policy,cederborg2015policy}, in our context we overload the term to refer to conditionally adjusting/perturbing actions at test time, without the privilege of policy refinement/update. Irrespective of the state coverage during training, deployed offline RL policies suffer from exogenous disturbances to transition dynamics. \callsign~ mitigates the impact of this exogenous process by characterizing the disturbances to shape the policy output.

The overall architecture of our proposed framework, \callsign~agents, is depicted in \autoref{fig:methodoverview}. The first two modules of the framework represent the offline training phase. Module 1 trains the agent using limited offline, mostly clean and disturbance-free, samples and learns the policy, $\pi^*_\mathcal{D}$. It also persists the value function $Q^*_\mathcal{D}$ which will be important later. Additionally, an autoencoder module 2 is trained using the offline trajectories by minimizing the reconstruction loss scaled by a clustering quality metric described later. The encoder-decoder framework captures the trend in the offline samples and learns the general trend in the data in some latent space, aiding in identifying disturbances and out-of-distribution (OOD) samples. The adaptation strategy of \callsign~agents during post-deployment is illustrated in module 3 (which also exploits the trained module 2). The reconstruction loss from the OOD detector serves as an identifier for OOD samples. The framework modifies the action space from the trained network and perturbs it in the direction of maximal increase in the Q function, with the amount of perturbation controlled by the OOD indicator, reconstruction loss. 

\subsection{Offline RL Training}
In module \circledfill{1}, while training policies from a dataset $\mathcal{D}$, we use existing model-free offline RL algorithms that use an actor-critic architecture to learn policies. Our method does not make any changes to the policy learning algorithm.

We use model-free offline RL algorithms such as Implicit Q-learning (IQL) to train policies from a dataset $\mathcal{D}$. IQL \cite{kostrikov2022offline} avoids distributional shifts during policy training by aiming to fit a Bellman optimal value function $Q^*$ without querying actions that are not supported by the data distribution. It introduces an additional critic for the state value function, and optimizes the value functions with the following losses:
\begin{equation}
    \min_{Q} \mathcal{L}^Q_{\text{IQL}}(Q) = \mathbb{E}_{(s,a,s')\sim\mathcal{D}}[(r(s,a) + \gamma V(s') - Q(s,a))^2],
\end{equation}
\begin{equation}
    \min_{V} \mathcal{L}^V_{\text{IQL}}(V) = \mathbb{E}_{(s,a)\sim\mathcal{D}}[L^2_\tau(\bar{Q}(s,a) - V(s))],
\end{equation}

where $L^2_\tau(u) = |\tau - \mathds{1}(u < 0)|u^2$ is the expectile loss with an expectile parameter $\tau$.

To extract the policy from the value function, we use an objective combining deep deterministic policy gradient and behaviour cloning (DDPG+BC \cite{fujimoto2021minimalist}), defined as
\begin{equation}
    \max_{\pi} \mathcal{J}_{\text{DDPG+BC}}(\pi) = \mathbb{E}_{s,a\sim\mathcal{D}}[Q(s, \mu^\pi(s)) + \alpha \log \pi(a | s)],
\end{equation}

where $\mu^\pi(s) = \mathbb{E}_{a\sim\pi(.|s)}[a]$ and $\alpha$ is a hyperparameter that controls the strength of the behaviour cloning regularizer.

\subsection{OOD Detector} 

We design the OOD or disturbance detector in module \circledfill{2}.
To learn the distribution of transition dynamics for sequences in trajectories, we fit an LSTM Autoencoder (LSTM-AE) on the sequences $\{(s_i^m, a_i^m, s^{m+1}_i, a^{m+1}_i, ... , a^{m+k}_i)_{i=1}^N\}$ present in the dataset $\mathcal{D}$, where $k$ is a hyperparameter that sets the sequence length.

The training objective of the LSTM\_AE is to minimize the reconstruction error. We introduce an additional guiding objective of learning the latent space on a manifold where trajectories are maximally separable. The loss function used for computing the reconstruction error is the Mean Squared Error (MSE), defined as $\sum_{i=1}^{D}(x^{inp}_i-x^{recon}_i)^2$, where $x^{inp}$ and $x^{recon}$ signify original input signal and reconstructed signals respectively by LSTM\_AE.
To guide the separation in the latent space, we perform k-means clustering on the encodings at the bottleneck layer, and estimate cluster separation using Davies-Bouldin score \cite{dbscore} 
\begin{equation}
\label{eq:DB}
    Score_{DB} = \max (\sum_{i} pairwise\mathbb{D}(X,c_i) / pairwise\mathbb{D}(c_i,c_j))
\end{equation}where $c_k$ is the centroid of cluster $k$. The final training objective is to minimize the reconstruction MSE loss scaled with Davies-Bouldin score.
\begin{equation}
\label{eq:finalloss}
    \ell_{LSTM\_AE} = \sum_{i=1}^{D}(x_i-x^\prime_i)^2 \otimes Score_{DB}
\end{equation}

Post-deployment, the LSTM-AE serves as an OOD detector and characterizer. It tries to reconstruct the most recent sequence while the agent is acting in the environment, and returns some reconstruction error. The reconstruction loss is a valid proxy for the metric highlighted in Equation~\ref{eq:probsetting}, $\rho_\exop = \left(s_\nu^{(input)}-s_\nu^{(recon)}\right)^2$ and it captures the position, span and frequency of the exogenous disturbance in the environment.  It is later used to learn the shaping potential $\phi$ that was defined earlier in \ref{sec:setting}.

We also explore alternative shallow detectors like isolation forests that give competitive results when compared against autoencoder-based approaches. However, these methods are not easily serializable to ONNX \cite{ONNX} models for deployment, and in some cases exhibit high variance in their predictions. Thus, our work focuses on methods that can easily be deployed on production systems.

\subsection{Post Deployment Policy Shaping}
Finally in module \circledfill{3} we outline the key aspect of \callsign~, the post-deployment action/policy shaping. 
To design the shaping potential $\phi(.|\pi*_\mathcal{D},\rho_\exop)$, we first analyze the problems with the policy. During training, the policy extraction regularizes the policy $\pi(a | s)$ with a behaviour cloning loss. This guides the policy to avoid OOD subspaces, but it also makes the policy's extrapolation poor when it encounters OOD regions, causing it to perform poorly \cite{Ackermann2024OfflineRL}. To shape the policy in these regions, we use the $Q$-function, which extrapolates better in these unknown subspaces. Specifically, we take the gradient of the $Q$-function with respect to $a\sim\pi^*_\mathcal{D}$, which is defined as $\nabla_a Q(s, a)$. This is likely to point towards actions with a higher value than those sampled from the policy \cite{park2024valuelearningreallymain}.



The value function gradient $\nabla_a Q(s, a)$ guides under disturbances in $S_\exop$, and the OOD metric $\rho_\exop$ captures the disturbances $S_\exop$ and $T_\exop$ acting on the environment. The shaping potential is designed by combining these signals, either using a connectionist model or an analytical model based on the domain and the data.

The shaping potential $\phi(.|\pi_\mathcal{D*},\rho_\exop)$ therefore becomes, $ \rho_\exop \otimes \nabla_a Q(s, a)$. 
The action sampled from the policy is adjusted as
\begin{equation}
    a^\prime = a\sim \pi^*_\mathcal{D} \oplus \alpha.k\sim \phi(\cdot|\pi^*_\mathcal{D},\rho_\exop),
\end{equation}

where $\alpha$ is a scaling parameter for numerical stability.


Action $a^\prime$ conditioned on the shaping potential $\phi(.|\pi*_\mathcal{D},\exop)$ approximates the post-deployment globally optimal action $a_\nu$.

\noindent \paragraph{Intuition}: The intuition behind the structure of the shaping potential is that, as also noted in our empirical evaluations, the two terms $\rho_\exop$ and $\nabla_a Q(s,a)$ contribute to two different aspects of the disturbance. Exogenous process $\exop$ may lead to both OOD $S_\exop$ and $T_\exop$. $\nabla_a Q(s,a)$ was learned on offline data and cannot detect or measure $T_\exop(s^\prime|s,a)$ if $s',s \in S_\mathcal{D}$. However if $s',s \in S_\exop$ then $\nabla_a Q(s,a)$ will manifest a non-zero gradient. $\rho_\exop$, on the other hand, is computed via the reconstruction loss of the LSTM\_AE, which are trained on transition sequences and change in transition dynamics with affect the sequence reconstruction fidelity. Thus $\rho_\exop$ approximately quantifies the position, span and amplitude of $T_\exop$ (which can also happen due to existence of $S_\exop$). $\otimes$ signify any generic combination function be it connectionist models or analytical formulae making  $\phi$ a good shaping potential. 

\subsection{Algorithms}
\begin{algorithm}[h]
\caption{Disturbance detector training}
\label{alg:OODtrain}
    \begin{algorithmic}[1]
    \Function{TrainDetector}{Offline Data $\mathcal{D}$}
    \State Initialize LSTM\_AE,  $\mathfrak{O}^\theta \gets [\mathfrak{O}_{encoder}, \mathfrak{O}_{decoder}]$
    \State Create sub-trajectories of length n; $\mathbf{X} \gets \tau^{(t-n, \ldots t)} \sim \mathcal{D}$
    \For{Minibatch $\{\tau\}_m \in \mathbf{X}$}
    \State Bottleneck embedding $\lbrace\langle emb\rangle\rbrace_m \gets \mathfrak{O}_{encoder}(\{\tau\}_m)$
    \State Clusters $\mathbb{C} \gets$ \Call{K-Means}{$\lbrace\langle emb\rangle\rbrace_m$} 
    \State $p \gets Score_{DB}(\mathbb{C})$ \Comment{from Eqn~\ref{eq:DB}}
    \State Reconstructed $\{\tau'\}_m \gets \mathfrak{O}_{decoder}(\lbrace\langle emb\rangle\rbrace_m)$
    \State Loss $\ell(\theta) = \left(\{\tau'\}_m-\{\tau\}_m\right)^2 \otimes p$ \Comment{from Eqn~\ref{eq:finalloss}}
    \State Update $\theta' = \theta + \nabla \ell_\theta$
    \EndFor
    \Return{$\mathfrak{O}^\theta$}
    \EndFunction
    \end{algorithmic}
\end{algorithm}

\begin{algorithm}[h]
\caption{Post Deployment shaping in \callsign}
\label{alg:PDshaping}
    \begin{algorithmic}[1]
    \Require OOD Detector Model $\mathfrak{O}$
    \Function{PDShaping}{$\langle \pi^*_\mathcal{D}, Q^*_\mathcal{D}\rangle, s_t\in S_\nu$, $\tau_t = \langle s_{}$} 
    \State $\rho_\exop \gets \mathfrak{O}(\tau_t)$ \Comment{uses a trajectory of sequence length $n$}
    \State Candidate action $a_t \gets \pi^*_\mathcal{D}(s_t)$
    \State Value Grad $\nabla Q^*_t = \frac{d}{d a_t}Q^*_\mathcal{D}(s_t,a_t)$
    \State Potential $k_t =  \phi(\cdot|\nabla Q^*_t, \rho_\exop)$
    \State Shaped Action $\hat{a}_t = a_t \oplus \alpha.k_t$ \Comment{from Eqn~\ref{eq:probsetting}}\\
    \Return{$a'_t$}
    \EndFunction
    \end{algorithmic}
\end{algorithm}
In this section we present the formal algorithms for the 2 main components of our framework, (1) The OOD/Disturbance detection and quantification framework , how its trained and what its outputs are, in function \textsc{TrainDetector}() in Algorithm~\ref{alg:OODtrain} and (2) The Post Deployment action/policy shaping mechanism in \callsign~that helps infer modified actions $a' \approx a_\nu \sim \pi_\nu$ subject to ODD/Disturbance metric, in function \textsc{PDshaping} in Algorithm~\ref{alg:PDshaping}.

\subsection{Discussion on the stability of \callsign~ agents}
The modified action at each time t can be represented as 
$ a_t^{'} = a_t + \alpha \nabla_{a} Q(s_t, a_t) \cdot \rho_{\exop} $
. 
$\bar{V}(s)$ represents the Lyapunov function and $\bar{V}(s_t)$ measures the deviation from the equilibrium state, $s^*$ \cite{khalil2015nonlinear}. \\
$ \bar{V}(s_t) \geq 0$ and $ \bar{V}(s^*) =0$.
For stability, the expected value of function should decrease,
$ \mathbb{E}  [\Delta \bar{V}(s_t) ] = \mathbb{E}  [\bar{V}(s_{t+1})- \bar{V}(s_{t})] <0$. Using transition probability in the above expression, 

$ \mathbb{E}  [\Delta \bar{V}(s_t)|s_t, a_t^{'} ] =  \int_{s'} \bar{V}(s') p(s'|s_t,a_t^{'}) ds'-\bar{V}(s_t) $. 
Using Taylor's series expansion around $a_t^{'}$ and by using the expression for the action, we get the following, 

\begin{eqnarray}
\mathbb{E}  [\Delta \bar{V}(s_t)|s_t, a_t^{'} ] \approx  \int_{s'} \bar{V}(s') p(s'|s_t,a_t) ds'-\bar{V}(s_t)  + \nonumber 
\\
\alpha \nabla_{a} Q(s_t, a_t) \cdot \rho_{\exop} \cdot \int_{s'} \bar{V}(s') \nabla_{a}p(s'|s_t,a_t) ds' \nonumber
\end{eqnarray}

For the system to be stable, $ \mathbb{E}  [\Delta \bar{V}(s_t)|s_t, a_t^{'} ] < 0 $, which results in the following condition, 

\begin{eqnarray}
\alpha \left| \nabla_{a} Q(s_t, a_t) \cdot \rho_{\exop} \cdot \int_{s'} \bar{V}(s') \nabla_{a}p(s'|s_t,a_t) ds'\right| \nonumber 
\\
< \left|\bar{V}(s_t) -\int_{s'} \bar{V}(s') p(s'|s_t,a_t) ds' \right| \nonumber
\end{eqnarray}
As long as the term, $\alpha \nabla_{a} Q(s_t, a_t) \cdot \rho_{\exop} $ remains bounded, the system behavior remains stable under the proposed policy. Intuitively,  as long as the  exogenous disturbance remains bounded, \callsign ~ maintains  stable behavior.  Deriving the exact conditions on the manifold for the exogenous disturbance with respect to the normal behavior is future work.

\begin{table*}[t]
\centering
\caption{Results Comparison of \callsign~vs baselines (Highest Scores in Bold)}
\vspace{-1em}
\label{tab:streetwise-results}
\scalebox{1.0}{
\begin{tabular}{lrrrrrrrr}
\hline
Experiment & UKF & IQL & IQL+OPEX & \callsign~ & \multicolumn{3}{c}{Gain \% of \callsign} \\
\cline{6-8}
 &  &  &  &  & from UKF & from IQL & from OPEX \\
\hline
RandomLoss\_1M\_L2 & 2.6644 & 2.5861 & 2.8382 & \textbf{3.0509} & 14.5060 & 17.9733 & 7.4940 \\
RandomLoss\_1M\_L1 & 2.9070 & 2.8625 & 3.0621 & \textbf{3.0918} & 6.3577 & 8.0104 & 0.9699 \\
FBw\_1 & 2.3378 & 2.5299 & 2.7104 & \textbf{2.7481} & 17.5538 & 8.6248 & 1.3909 \\
FBw\_2 & 1.9311 & 1.9733 & 1.9001 & \textbf{2.0187} & 4.5340 & 2.3007 & 6.2417 \\
BL\_100k\_L25 & 1.1109 & 1.1086 & 1.0986 & \textbf{1.1368} & 2.3310 & 2.5396 & 3.4747 \\
BL\_1M\_L25 & 1.9987 & 2.3700 & 2.2758 & \textbf{2.3706} & 18.6064 & 0.0000 & 4.1655 \\
FBL\_300kbps & \textbf{1.4226} & 1.3747 & 1.3899 & 1.4099 & -0.8927 & 2.5605 & 1.4389 \\
FBL\_500kbps & 1.8609 & 1.9135 & 1.9484 & \textbf{1.9533} & 4.9689 & 2.0799 & 0.2514 \\
Stable\_1M & 3.2038 & 3.1861 & 3.5710 & \textbf{3.6004} & 12.3700 & 13.0030 & 0.8232 \\
Stable\_3M & \textbf{4.1921} & 3.9593 & 4.1632 & 4.1627 & -0.7013 & 5.1372 & -0.0120 \\
Stable\_4M & 4.2245 & 4.0239 & 4.2042 & \textbf{4.2302} & 0.5700 & 5.1268 & 0.6184 \\
\hline
\end{tabular}
}
\end{table*}


\begin{figure*}[t]
    \centering
    \scalebox{0.38}{\input{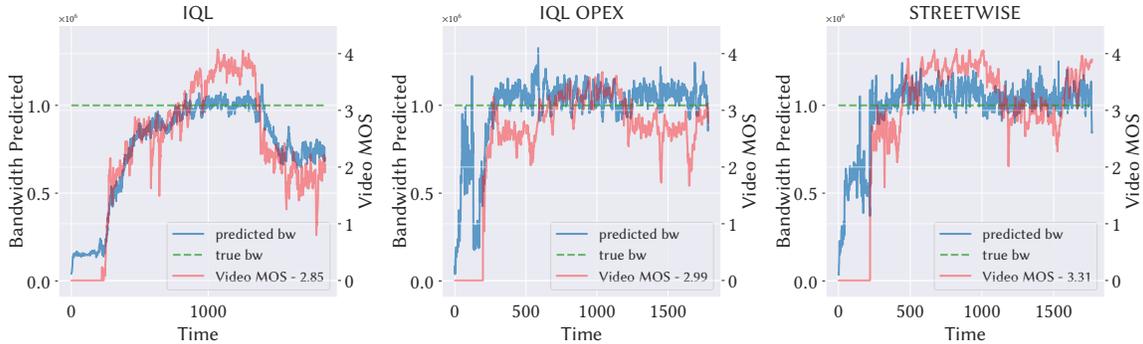}}
    \vspace{-1em}
    \caption{Comparison of model predicted bandwidths and video MOS on calls with the Random Loss network profile.}
    \vspace{-1em}
  \label{fig:cifx}
\end{figure*}

\section{Evaluation}
\label{sec:eval}
We evaluate the effectiveness of \callsign~based on the following research questions: 
\begin{enumerate}
    \item[Q1.] How effective is \callsign~in adaptively handling intermittent exogenous disturbances post-deployment in real Bandwidth Estimation task compared to baselines?
    \item[Q2.] Is \callsign~ generalizable to other domains? 
    
\end{enumerate}

\subsection{Experiment Domains}

\subsubsection{Primary: Bandwidth Estimation in RTC}
We choose Microsoft's Bandwidth Estimation Challenge \cite{challengePaper} as the platform to run our experiments on. The objective of the challenge is to learn policies that maximise QoE on RTC systems using offline RL techniques. The challenge provides a dataset of real-world trajectories collected from audio/video peer-to-peer (P2P) Microsoft Teams calls. One dataset is collected from calls that are conducted on test bed nodes geographically distributed across many countries, and the other dataset is collected from calls conducted between pairs of machine connected via a network emulation software.

The trajectories are collected by multiple behaviour policies, such as WebRTC, Kalman-filtering-based methods and other ML policies. For a sequence $(o_n, a_n, r_n^{audio}, r_n^{video})$ in a given call trajectory, $o_n$ is a 150-dimensional vector computed from packet information received by the client, $a_n$ is the bandwidth predicted by the estimator (behaviour policy) during the call, and the rewards $r_n^{audio}, r_n^{video}$ are provided in the form of objective audio quality and video quality signals that predict user-perceived quality \cite{mos_model}. 


The models are evaluated on an emulation platform that conducts P2P test calls across diverse network scenarios. This platform comprises of a network of interconnected nodes that simulate various link conditions. It replicates predefined network traces, ensuring consistent test environments. This setup enables systematic model evaluation across diverse simulated network scenarios.

\subsubsection{Other Domains: MuJoCo Disturbed Dynamics Continuous Control Tasks}
Since there are no existing benchmarks that evaluate the performance of an agent under transition dynamics disturbances at test time, we simulate it by adding stochastic noise to environment parameters of MuJoCo environments. We conduct our evaluations on three MuJoCo Gym locomotion tasks: HalfCheetah, Hopper and Walker2D, since their datasets are provided in the D4RL benchmark.

\subsection{Experimental Settings}

\subsubsection{Baselines}

We compare \callsign~ against the following baselines: (1) Unscented Kalman Filter (UKF) (2) Implicit Q-Learning (IQL) \cite{kostrikov2022offline} (3) OPEX + IQL \cite{park2024valuelearningreallymain}.

UKF with a rule-based controller is a reference bandwidth estimator that has been deployed in production. Most recent works cite IQL as a strong baseline for dealing with OOD subspaces, hence it is our primary baseline. OPEX is a test-time policy extraction strategy that shows better results than base IQL, hence we choose it as another baseline.

We are aware of works that use diffusion policies \cite{zhou2023adaptiveonlinereplanningdiffusion, wang2023diffusion, Ada_2024} to deal with OOD, but we do not implement them because they do not adhere to the compute constraints of the RTC domain.

While robust policies can be trained using online RL, to do so, they have to be trained on various noisy profiles with appropriate feedback. In real environments, this can be done using pre-production test environments with similar traffic. However, we have set the context such that we only have access to a limited set of data that we train offline RL policies from. Thus, we do not use online RL policies as baselines.

\subsubsection{Evaluation Metrics}

(1) {Bandwidth Estimation}. The evaluation criteria for the challenge models is the user-perceived quality, which is represented by the {audio MOS} and {video MOS} (Mean Opinion Score). \cite{ITU-T_P.mos1, ITU-T_P.mos2}. Both audio and video MOS values are computed on a scale of $[0, 5]$. 
(2) {MuJoCo Tasks}. For MuJoCo-Gym experiments, the continuous control locomotion tasks have their reward defined as the distance the agent covers along the x-axis during the episode.

\subsubsection{Implementation}
Our implementation choices prioritize efficiency in the models. For the bandwidth estimator, we implement the actor and critic networks as two-layer MLPs with 128 hidden parameters each and Tanh activations. The models used in the MuJoCo experiments have similar architectures, with their hidden layers having 256 parameters instead to follow the original IQL implementation \cite{kostrikov2022offline}.

The OOD detector is implemented as an LSTM autoencoder that takes as input the last $k$ state-action transitions, structured as state-action-next-state and so on, and reconstructs it. The optimum sequence length $k$ depends on the model architecture and distribution of the data. For our experiments, we set $k$ to 5.
The autoencoder has two LSTM units each in the encoder and decoder. The hidden layers have 32 parameters, and the bottleneck layer is encoded to a vector of length 16. The reconstruction loss of the autoencoder $\rho_\exop$ is the proxy for the metric highlighted in Equation~\ref{eq:probsetting}.


\subsubsection{Noise in MuJoCo environments}
\label{sec:mujoconoise}
In our experiments, we introduce custom noise to the MuJoCo environment by dynamically varying the viscosity throughout an episode using the viscosity attribute of the environment. This function simulates non-stationary dynamics by introducing periodic spikes and plateaus in the viscosity, forcing the agent to adapt to fluctuating environmental conditions. The viscosity begins with a base value and is altered by multiple randomly generated spikes, where each spike’s amplitude and direction (positive or negative) are randomly determined. The duration of both the spike and the following plateau is parameterized to control how long the viscosity stays elevated or reduced. To maintain physical plausibility, the viscosity is constrained to stay above a minimum threshold (10\% of the base value).

By adding this variability, the environment becomes more challenging, as sudden increases in viscosity directly impair the agent's ability to move efficiently, making actions less effective. This dynamic resistance reduces the expected return since the agent struggles to maintain optimal trajectories under the fluctuating conditions. The noise models real-world uncertainties, such as environmental friction or fluid resistance, providing a more rigorous test for the robustness of our approach.

\subsection{Experimental Results}

\subsubsection{Bandwidth Estimation}

To validate our approach, we compare its performance against the aforementioned baselines on 10+ network profiles. For each profile, we conduct 30 calls with each of the baseline estimator models and our estimator model, and then report the average score across these calls. 
Network profiles can be characterized by their bandwidth patterns and packet loss patterns. Packet loss in calls can occur either randomly or in bursts. Profiles with a fluctuating bandwidth typically exhibit periodic or non-periodic oscillations in the bandwidth.
Thus, our evaluation network profiles can be broadly classified into 5 categories: 
\begin{enumerate}
    \item Random Loss: Packets are dropped unpredictably at random intervals throughout data transmission.
    \item Burst Loss (BL): Multiple consecutive packets are lost in short periods, creating "bursts" of data loss.
    \item Fluctuating Bandwidth (FBw): The maximum rate of data transmission varies over time, increasing and decreasing periodically.
    \item Fluctuating Burst Loss (FBL): Both the occurrence and intensity of burst losses change over time, creating a highly variable network behavior.
    \item Stable Bandwidth: The network maintains a constant data transmission rate without unexpected fluctuations.
\end{enumerate}

For our evaluations, we select 2 profiles in each noisy category, and 3 profiles with stable bandwidths. Some profile names indicate what bandwidth segment they belong to. Profiles with loss indicate the level of packet loss present. In our results, we only report the video MOS, since there is negligible headroom for improvement in the audio MOS when compared to the baselines.


\textbf{Aggregate Effectiveness.} Results in \autoref{tab:streetwise-results} show the highest gains in the random loss profile. \callsign~ shows upto $\sim$18\% improvement in the video MOS. IQL+OPEX also performs well, showing $\sim 1 - 7.5$\% gains in the video MOS. This adaptation to random packet loss shows that shaping policies post-deployment can be effective. 
Interestingly, burst loss calls show a different trend in the results when compared to random loss calls. IQL+OPEX performs worse than just the IQL policy, while \callsign~ shows similar or slightly similar scores. We note that IQL+OPEX fails to adapt to a burst of packet drops, but \callsign~ maintains performance equal to or better than the other baseline policies. Note that on an extremely stable call like ``3M'', UKF does slightly better. The likely cause is that UKF is a smooth predictor whereas Offline RL policies have some associated variance, which may at times cause aggregate returns to slightly diminish.

\textbf{In call analysis.} The predicted bandwidth of the models and the video MOS throughout the calls, conducted on a profile with random loss, are shown in \autoref{fig:cifx}.
IQL has a noticeable difference in prediction performance compared to the other two models. IQL+OPEX and \callsign~ estimate the bandwidth much more accurately, adapting especially well to tricky segments such as the beginning of the call and the drop in bandwidth midway in the call. 

The results also suggest that our model performs reliably on stable profiles, and when compared to the baselines, our post-deployment shaping potential is still guiding the policy to towards actions with higher returns. This affirmatively answers \textbf{(Q1)}.

\subsubsection{MuJoCo Disturbed Continuous Control Tasks}

\begin{table}[h!]
\centering
\caption{Results on Mujoco env. with exogenous noise}
\vspace{-.5em}
\label{tab:mujoco-results}
\scalebox{0.7}{
\begin{tabular}{llcccc}
\hline
Difficulty & Env & IQL & \multicolumn{2}{c}{IQL + OPEX} & \callsign~ \\
\cline{4-5}
& & & $\beta=0.1$ & $\beta=0.01$ & \\
\hline
\multirow{3}{*}{\rotatebox[origin=c]{90}{Medium}} & HalfCheetah & 23.22 $\pm$ 2.57 & 23.98 $\pm$ 2.52 & 22.58 $\pm$ 2.35 & 24.67 $\pm$ 2.84 \\
& Hopper & 85.29 $\pm$ 2.34 & 86.01 $\pm$ 2.23 & 84.60 $\pm$ 2.19 & 86.01 $\pm$ 2.36 \\
& Walker2d & 65.57 $\pm$ 2.38 & 66.48 $\pm$ 2.49 & 66.31 $\pm$ 2.42 & 66.37 $\pm$ 2.63 \\
\hline
\multirow{3}{*}{\rotatebox[origin=c]{90}{Expert}} & HalfCheetah & 30.60 $\pm$ 3.82 & 8.45 $\pm$ 5.51 & 30.81 $\pm$ 3.87 & 30.63 $\pm$ 3.73 \\
& Hopper & 68.85 $\pm$ 32.84 & 47.28 $\pm$ 24.50 & 45.13 $\pm$ 21.49 & 60.30 $\pm$ 27.65 \\
& Walker2d & 84.82 $\pm$ 4.37 & 86.75 $\pm$ 3.28 & 85.31 $\pm$ 4.28 & 86.25 $\pm$ 7.74 \\
\hline
\multirow{3}{*}{\rotatebox[origin=c]{90}{Random}} & HalfCheetah & 2.60 $\pm$ 0.65 & 15.71 $\pm$ 0.97 & 9.18 $\pm$ 1.41 & 14.80 $\pm$ 0.81 \\
& Hopper & 9.04 $\pm$ 0.31 & 9.15 $\pm$ 0.42 & 9.22 $\pm$ 0.71 & 9.37 $\pm$ 0.40 \\
& Walker2d & 7.26 $\pm$ 0.41 & 8.84 $\pm$ 4.71 & 5.32 $\pm$ 0.16 & 7.14 $\pm$ 0.56 \\
\hline
\multirow{3}{*}{\rotatebox[origin=c]{90}{\parbox{10mm}{Medium-Expert}}} & HalfCheetah & 29.90 $\pm$ 3.63 & 18.69 $\pm$ 1.20 & 30.37 $\pm$ 3.60 & 27.99 $\pm$ 3.69 \\
& Hopper & 90.27 $\pm$ 3.21 & 70.02 $\pm$ 32.04 & 90.76 $\pm$ 3.65 & 88.95 $\pm$ 5.48 \\
& Walker2d & 71.42 $\pm$ 5.46 & 82.87 $\pm$ 7.05 & 70.72 $\pm$ 3.92 & 79.62 $\pm$ 8.24 \\
\hline
\multirow{3}{*}{\rotatebox[origin=c]{90}{\parbox{10mm}{Medium-Replay}}} & HalfCheetah & 21.57 $\pm$ 2.50 & 22.90 $\pm$ 2.31 & 22.20 $\pm$ 2.17 & 23.97 $\pm$ 2.78 \\
& Hopper & 86.98 $\pm$ 3.55 & 85.29 $\pm$ 2.91 & 86.20 $\pm$ 3.22 & 86.45 $\pm$ 3.33 \\
& Walker2d & 65.66 $\pm$ 6.70 & 68.40 $\pm$ 3.41 & 65.64 $\pm$ 2.87 & 68.20 $\pm$ 3.63 \\
\hline
\end{tabular}
}
\end{table}



To evaluate the generalization of \callsign~, we perform additional  experiments on the MuJoCo domain (results in \autoref{tab:mujoco-results}). To simulate exogenous disturbances, we train policies offline on datasets without noisy trajectories and deploy them on noise post-deployment environments (oultined in \autoref{sec:mujoconoise}). For any given environment, D4RL provides multiple datasets that are collected by a mixture of behaviour policies. We report the results for all the datasets provided for the MuJoCo Gym continuous control environments. IQL and IQL+OPEX as our baselines here. For OPEX, we report results for multiple values of the hyperparameter $\beta$. We present average scores over 100 episodes for each task.

The results on HalfCheetah shed light on some interesting insights. First, we note that \callsign~ performs better than IQL in most scenarios. Policies trained on non-expert datasets show the highest improvement when \callsign~ is implemented. 

Our second observation is that although OPEX gives the highest returns with the optimal choice of $\beta$, $\beta$ itself is very sensitive to the training dataset. Comparing the results of the Medium and Expert datasets, we see that the performance of OPEX strongly deteriorates for $\beta=0.1$. The impracticality of performing a line search for optimal $\beta$ in production deployments highlights an advantage of \callsign~, which demonstrates consistent performance across all tasks due to its dynamic disturbance adjustment capabilities. These observations affirmatively answer \textbf{(Q2)}.

Note that, in \textit{Hopper} and \textit{Walker2D}, disturbing viscosity physics had an unintended side-effect of non-stationary reward functions. Neither \callsign~nor OPEX are designed for non-stationary rewards and hence post-deployment perturbations do not always exhibit higher returns. OPEX just seems to be doing marginally better some instances where it is getting lucky due to scaling with high value of $\beta$, since these comparatively simple environments encourage numerically higher values of actions. Adding post-deployment support for disturbed reward functions is one of our future plans.



\textbf{Extended Evaluations.} We conducted additional experiments with varying gravity and wind speed to introduce environmental noise. Gravity variations included burst, sinusoidal, random, and step changes. However, gravity modifications proved unsuitable for robustness testing due to unintended effects on normal forces and task dynamics. Wind speed variations, primarily in the horizontal direction, utilized sinusoidal patterns with adjustable amplitudes to simulate gusts. This approach led to excessive agent acceleration or deceleration, causing fluctuating returns and skewed performance metrics. Ultimately, these experiments demonstrated that gravity and wind modifications excessively disrupted task dynamics, rendering them ineffective for evaluating our approach's robustness.




\section{Conclusion}

Our proposed, first of its kind, \callsign~agent is a significant step in the direction of building agents that are capable of enduring unseen dynamics changes post-deployment. Significant quality improvements on a sensitive domains like RTC demonstrate its effectiveness, while stability on standard benchmarks illustrate the generalizability of our method. We believe that shaping policies/actions post-deployment is an area that presents plenty of opportunities, and with our work, we seek to motivate more research in this direction. Adaptation to more real sensitive and safety-critical domains and providing some regret bounds and guarantees are some of our immediate future research directions. 

\bibliographystyle{ACM-Reference-Format}
\bibliography{arxiv}


\end{document}